\documentclass{article}

\usepackage[dblblindworkshop,final]{neurips_2026}  
\workshoptitle{Diffusion Language Models: Foundations, Efficiency, and Reasoning}

\usepackage[utf8]{inputenc}
\usepackage[T1]{fontenc}
\usepackage{amsmath,amssymb}
\usepackage{booktabs}
\usepackage{graphicx}
\usepackage{xcolor}
\usepackage{microtype}
\usepackage{tikz}
\usetikzlibrary{arrows.meta,decorations.pathreplacing,positioning}
\usepackage{pifont}
\usepackage{subcaption}
\usepackage{enumitem}
\usepackage{placeins}
\usepackage[hidelinks]{hyperref}

\definecolor{dcol}{HTML}{23479C}   
\definecolor{wcol}{HTML}{C4541D}   
\definecolor{tcol}{HTML}{0F7B72}   
\newcommand{\dc}[1]{\textcolor{dcol}{#1}}
\newcommand{\wc}[1]{\textcolor{wcol}{#1}}
\newcommand{\tc}[1]{\textcolor{tcol}{#1}}
\newcommand{\cnum}[1]{\textbf{\ding{\numexpr181+#1\relax}}}

\newcommand{\Nconfigs}{15}
\newcommand{\Nseeds}{3}

\newcommand{\NTgrid}{1, 2, 4, 8, 16}
\newcommand{\Nsurface}{75}
\newcommand{\Nruns}{45}
\newcommand{\Nitems}{400}

\newcommand{\Ngpuhours}{162.5}
\newcommand{\Nbaselr}{0.004}
\newcommand{\Nsecperchar}{0.0602}
\newcommand{\Ndtau}{0.110}
\newcommand{\Ndtauci}{[0.092,\,0.131]}
\newcommand{\Ntauwer}{0.836}
\newcommand{\Ntauwerci}{[0.825,\,0.851]}
\newcommand{\Ntausim}{0.726}
\newcommand{\Ntausimci}{[0.709,\,0.744]}
\newcommand{\Nkappawer}{3.000}

\newcommand{\Nkappasim}{3.000}

\newcommand{\NtauWER}{0.836}

\newcommand{\NalphaBWER}{1.574}
\newcommand{\NbetaBWER}{2.040}
\newcommand{\NaiccFNWER}{-15.5}
\newcommand{\NaiccSSWER}{69.3}

\newcommand{\NmapeWER}{20.2}
\newcommand{\NmapeNWER}{20.4}
\newcommand{\NtauSIM}{0.726}

\newcommand{\NalphaBSIM}{0.919}
\newcommand{\NbetaBSIM}{0.553}
\newcommand{\NaiccFNSIM}{-5.5}
\newcommand{\NaiccSSSIM}{71.3}

\newcommand{\NmapeSIM}{2.4}
\newcommand{\NmapeNSIM}{2.2}
\newcommand{\Nbootreps}{2000}
\newcommand{\Nseedlist}{{0, 1, 2}}
\newcommand{\Ngfour}{passed}

\newcommand{\Ndegentsixteen}{0.1}

\newcommand{\NmaxN}{133}
\newcommand{\NminN}{19}
\newcommand{\NdepthgainA}{40.2}
\newcommand{\NdepthgainB}{16.3}
\newcommand{\NdepthgainC}{7.3}

\newcommand{\NsimspreadTsixteen}{0.065}
\newcommand{\NabsWERgain}{0.952}
\newcommand{\NabsSIMgain}{0.161}
\newcommand{\NmagAsym}{5.9$\times$}

\newcommand{\Ndtaufour}{+0.1095}
\newcommand{\Ndtaufourci}{[0.0951, 0.1241]}
\newcommand{\NmaxNfour}{276}
\newcommand{\Ndtaunineok}{+0.1767}
\newcommand{\Ndtaunineokci}{[0.0789, 0.2854]}
\newcommand{\Ndtauasr}{+0.1283}
\newcommand{\Ndtaulogamp}{+0.1260}

\newcommand{\NwerTsixteenExt}{0.1516}
\newcommand{\NwerTsixtyfour}{0.1294}
\newcommand{\Nbudgetc}{125\,M}
\newcommand{\NnfeCoarseWER}{0.2191}
\newcommand{\NnfeCoarseSIM}{0.3057}
\newcommand{\NnfeCoarseUT}{2.17}
\newcommand{\NnfeCoarseDegen}{0.1}
\newcommand{\NnfeUniformWER}{0.3144}
\newcommand{\NnfeUniformSIM}{0.3477}
\newcommand{\NnfeUniformUT}{2.51}
\newcommand{\NnfeUniformDegen}{0.5}
\newcommand{\NnfeFineWER}{1.1480}
\newcommand{\NnfeFineSIM}{0.2246}
\newcommand{\NnfeFineUT}{1.56}
\newcommand{\NnfeFineDegen}{39.1}
\newcommand{\Nfivepointtwo}{5.2}

\newcommand{\NsimRoundtrip}{0.555}
\newcommand{\NsimRealAudio}{0.678}
\newcommand{\NcodecShare}{62}
\newcommand{\NsearchDelta}{+0.0444}
\newcommand{\NsearchCI}{[0.0416, 0.0470]}
\newcommand{\NsearchWin}{72}
\newcommand{\NsearchWER}{0.1955}
\newcommand{\NrefineWER}{0.1316}

\newcommand{\NrandomPick}{0.4259}

\newcommand{\NsearchVsDefault}{+0.0365}

\newcommand{\NsearchWERcost}{+0.0425}
\newcommand{\NselCorrRange}{0.71--0.73}
\newcommand{\NsearchD}{+0.0422}
\newcommand{\NsearchNinetyk}{+0.0368}

\newcommand{\NsearchMaxN}{276}
\newcommand{\NheadroomSim}{0.0747}
\newcommand{\NparamGainSim}{+0.0531}
\newcommand{\NtrainGainSim}{+0.0827}
\newcommand{\NtrainGainCI}{[0.0733, 0.0908]}
\newcommand{\NtrainGainRuns}{9}
\newcommand{\NtrainDouble}{+0.0155}
\newcommand{\NguideGainSim}{+0.0087}
\newcommand{\NguideWER}{+4.41}
\newcommand{\NrateGainSim}{+0.0044}
\newcommand{\NdtauAsrCI}{[0.1064, 0.1492]}
\newcommand{\NdtauLogCI}{[0.1043, 0.1474]}
\newcommand{\NallocCwer}{0.2191}
\newcommand{\NallocCsim}{0.3057}
\newcommand{\NallocCut}{2.17}
\newcommand{\NallocCdeg}{0.1}
\newcommand{\NallocUwer}{0.3144}
\newcommand{\NallocUsim}{0.3477}
\newcommand{\NallocUut}{2.51}
\newcommand{\NallocUdeg}{0.5}
\newcommand{\NallocFwer}{1.1480}
\newcommand{\NallocFsim}{0.2246}
\newcommand{\NallocFut}{1.56}
\newcommand{\NallocFdeg}{39.1}
\newcommand{\NsrchLod}{+0.0153}
\newcommand{\NsrchLoci}{[0.0127, 0.0179]}

\newcommand{\NsrchLorw}{0.1530}
\newcommand{\NsrchLosw}{0.1938}
\newcommand{\NsrchLoT}{16}
\newcommand{\NsrchLoK}{2}
\newcommand{\NsrchMidd}{+0.0322}
\newcommand{\NsrchMidci}{[0.0307, 0.0338]}

\newcommand{\NsrchMidrw}{0.1335}
\newcommand{\NsrchMidsw}{0.1947}
\newcommand{\NsrchMidT}{32}
\newcommand{\NsrchMidK}{4}
\newcommand{\NsrchHid}{+0.0444}
\newcommand{\NsrchHici}{[0.0416, 0.0470]}

\newcommand{\NsrchHirw}{0.1316}
\newcommand{\NsrchHisw}{0.1955}
\newcommand{\NsrchHiT}{64}
\newcommand{\NsrchHiK}{8}

\newcommand{\NscopeDd}{+0.0422}
\newcommand{\NscopeDci}{[0.0393, 0.0448]}

\newcommand{\NscopeNd}{+0.0368}
\newcommand{\NscopeNci}{[0.0321, 0.0428]}

\newcommand{\NctxWerCost}{71.7}
\newcommand{\NctxAsim}{0.3213}
\newcommand{\NctxAwer}{0.1578}
\newcommand{\NctxBsim}{0.3839}
\newcommand{\NctxBwer}{0.1474}
\newcommand{\NctxCsim}{0.3724}
\newcommand{\NctxCwer}{0.7692}
\newcommand{\NctxDsim}{0.3759}
\newcommand{\NctxDwer}{0.8642}
\newcommand{\NsimThirtyk}{0.4081}
\newcommand{\NsimNinetyk}{0.4807}
\newcommand{\NsimOneeightyk}{0.4962}
\newcommand{\NtrainDoubleCI}{[0.0064, 0.0243]}
\newcommand{\NtrainResid}{-0.0302}
\newcommand{\NtrainPred}{0.5264}
\newcommand{\NgamLosim}{+0.0087}
\newcommand{\NgamLoci}{[0.0063, 0.0115]}
\newcommand{\NgamLower}{+4.41}

\newcommand{\NgamMidsim}{-0.0021}
\newcommand{\NgamMidci}{[-0.0059, 0.0023]}
\newcommand{\NgamMidwer}{+8.94}

\newcommand{\NgamHisim}{-0.0322}
\newcommand{\NgamHici}{[-0.0367, -0.0274]}
\newcommand{\NgamHiwer}{+19.10}

\newcommand{\NrateCI}{[0.0017, 0.0067]}
\newcommand{\NrateRuns}{13}
\newcommand{\NrateP}{0.051}
\newcommand{\NrateWER}{+0.86}
\newcommand{\Nclayerrange}{0.493--0.514}

\newcommand{\NtrainHours}{2{,}000}
\newcommand{\NtrainSpk}{4{,}000}
\newcommand{\NheldoutSpk}{200}
\newcommand{\NevalItems}{400}
\newcommand{\NevalSpeakers}{174}
\newcommand{\NwerFloor}{0.0345}
\newcommand{\NsimCeiling}{0.5554}
\newcommand{\NgapWER}{86.2}
\newcommand{\NgapWERci}{[82.8, 89.1]}
\newcommand{\NgapSIM}{46.4}
\newcommand{\NgapSIMci}{[45.0, 47.8]}
\newcommand{\NgapRatio}{1.86$\times$}
\newcommand{\NgapRatioci}{[1.82, 1.90]}
\newcommand{\NgapRatioLo}{1.68$\times$}
\newcommand{\NgapRatioHi}{1.86$\times$}
\newcommand{\NgapWERext}{88.4}
\newcommand{\NgapSIMext}{47.9}
\newcommand{\NgapSIMextci}{[46.6, 49.2]}

\newcommand{\NgapRatioTLo}{1.85$\times$}
\newcommand{\NgapRatioTHi}{1.96$\times$}
\newcommand{\NgapTmax}{64}
\newcommand{\NdtauNflip}{3}
\newcommand{\NdtauRangeSixteen}{+0.1102}
\newcommand{\NdtauRangeSixtyfour}{+0.0660}

\newcommand{\NdtauRangeEight}{+0.2191}
\newcommand{\NdtauHonestci}{[0.0349, 0.1640]}
\newcommand{\NchiWER}{14.3}
\newcommand{\NchiSIM}{8.8}
\newcommand{\NkappaImpliedWER}{0.41}

\newcommand{\NTmaxWERdelta}{+0.0018}
\newcommand{\NTmaxSIMdelta}{-0.0052}

\newcommand{\NmencNfam}{4}
\newcommand{\NmencLo}{64.6}
\newcommand{\NmencHi}{79.0}
\newcommand{\NmencRuns}{15}
\newcommand{\NmencItems}{200}
\newcommand{\NmencMinAuc}{0.98}

\newcommand{\NgapThirtykRatio}{1.84$\times$}

\newcommand{\NgapNinetykRatio}{1.36$\times$}

\newcommand{\NgapEightykRatio}{1.23$\times$}

\newcommand{\NsatWER}{-0.09}
\newcommand{\NsatWERasrTwo}{-0.06}
\newcommand{\NsatSIM}{+7.4}
\newcommand{\NasrTwoName}{wav2vec2-large-960h-lv60-self}
\newcommand{\NasrTwoFloor}{0.1501}
\newcommand{\NtrendRuns}{9}
\newcommand{\NtrendBudgets}{3}

\title{Refinement Buys Intelligibility, Search Buys Identity:\\
What Test-Time Compute Buys in Masked-Diffusion TTS}
\definecolor{tvink}{HTML}{1C2420}
\definecolor{tvpaper}{HTML}{F5F5EF}
\newcommand{\tensorvizmark}[1][1.25em]{%
  \raisebox{-0.27em}{\begin{tikzpicture}[x=#1/32,y=-#1/32,line cap=round,line width=0.07*#1]
    \draw[tvink] (8,7)--(24,7) (8,7)--(8,25) (8,16)--(24,16) (24,7)--(24,25) (8,25)--(24,25);
    \foreach \p in {(8,7),(24,7),(8,25),(24,25),(24,16)}
      \filldraw[draw=tvink,fill=tvpaper] \p circle (3);
  \end{tikzpicture}}}
\newcommand{\tensorvizname}{{\sffamily\fontseries{b}\selectfont tensor}{\sffamily\fontseries{m}\selectfont Viz}}
\newcommand{\tensorviz}{\mbox{\tensorvizmark\hspace{0.4em}\tensorvizname}}

\usetikzlibrary{svg.path}
\newcommand{\githubicon}{\raisebox{-0.12em}{\resizebox{!}{0.95em}{\tikz[yscale=-1]\fill svg{M12 .297c-6.63 0-12 5.373-12 12 0 5.303 3.438 9.8 8.205 11.385.6.113.82-.258.82-.577 0-.285-.01-1.04-.015-2.04-3.338.724-4.042-1.61-4.042-1.61C4.422 18.07 3.633 17.7 3.633 17.7c-1.087-.744.084-.729.084-.729 1.205.084 1.838 1.236 1.838 1.236 1.07 1.835 2.809 1.305 3.495.998.108-.776.417-1.305.76-1.605-2.665-.3-5.466-1.332-5.466-5.93 0-1.31.465-2.38 1.235-3.22-.135-.303-.54-1.523.105-3.176 0 0 1.005-.322 3.3 1.23.96-.267 1.98-.399 3-.405 1.02.006 2.04.138 3 .405 2.28-1.552 3.285-1.23 3.285-1.23.645 1.653.24 2.873.12 3.176.765.84 1.23 1.91 1.23 3.22 0 4.61-2.805 5.625-5.475 5.92.42.36.81 1.096.81 2.22 0 1.606-.015 2.896-.015 3.286 0 .315.21.69.825.57C20.565 22.092 24 17.592 24 12.297c0-6.627-5.373-12-12-12};}}}
\newcommand{\hficon}{\raisebox{-0.12em}{\resizebox{!}{0.95em}{\tikz[yscale=-1]\fill svg{M1.4446 11.5059c0 1.1021.1673 2.1585.4847 3.1563-.0378-.0028-.0691-.0058-.1058-.0058-.4209 0-.8015.16-1.0704.4512-.3454.3737-.4984.8335-.4316 1.293a1.576 1.576 0 0 0 .2148.5978c-.2319.1864-.4018.4456-.4844.7578-.0646.2448-.131.7543.2149 1.2794a1.4552 1.4552 0 0 0-.0625.1055c-.208.3923-.2207.8372-.0371 1.25.2783.6258.9696 1.1175 2.3126 1.6467.8356.3292 1.5988.5411 1.6056.543 1.1046.2847 2.104.4277 2.969.4277 1.4173 0 2.4754-.3849 3.1525-1.1446 1.538.2651 2.791.1403 3.592.006.6773.7555 1.7332 1.1387 3.1467 1.1387.8649 0 1.8643-.143 2.969-.4278.0068-.0019.77-.2138 1.6056-.543 1.343-.5292 2.0343-1.0208 2.3126-1.6466.1836-.4129.171-.8577-.037-1.25a1.4685 1.4685 0 0 0-.0626-.1056c.346-.525.2795-1.0346.2149-1.2793-.0826-.3122-.2525-.5714-.4844-.7579.11-.1816.1831-.3788.2148-.5977.0669-.4595-.0862-.9193-.4316-1.293-.2688-.2913-.6495-.4513-1.0704-.4513-.0209 0-.0376.0008-.0588.0018.3162-.9966.4846-2.0518.4846-3.1523 0-5.807-4.7362-10.5144-10.5789-10.5144-5.8426 0-10.5788 4.7073-10.5788 10.5144Zm10.5788-9.4831c5.2727 0 9.5476 4.246 9.5476 9.483a9.4201 9.4201 0 0 1-.2696 2.2365c-.0039-.0047-.0079-.011-.0117-.0156-.274-.3255-.6679-.5059-1.1075-.5059-.352 0-.714.1155-1.0763.3438-.2403.1517-.5058.422-.7793.7598-.2534-.3492-.608-.5832-1.0137-.6465a1.5174 1.5174 0 0 0-.2344-.0176c-.9263 0-1.4828.7993-1.6935 1.5177-.1046.2426-.6065 1.3482-1.3614 2.0978-1.1681 1.1601-1.4458 2.3534-.8396 3.6382-.843.1029-1.5836.0927-2.365-.006.5906-1.212.3626-2.4388-.8426-3.6322-.755-.7496-1.2568-1.8552-1.3614-2.0978-.2107-.7184-.7673-1.5177-1.6935-1.5177-.078 0-.1568.0054-.2344.0176-.4057.0633-.7604.2973-1.0137.6465-.2735-.3379-.539-.6081-.7794-.7598-.3622-.2283-.7243-.3438-1.0762-.3438-.4266 0-.8094.171-1.0821.4786a9.4208 9.4208 0 0 1-.2598-2.1936c0-5.237 4.2749-9.483 9.5475-9.483zM8.6443 7.0036c-.4838.0043-.9503.2667-1.1934.7227-.3536.6633-.1006 1.4873.5645 1.84.351.1862.4883-.5261.836-.6485.3107-.1095.841.399 1.0078.086.3536-.6634.1025-1.4874-.5625-1.84a1.3659 1.3659 0 0 0-.6524-.1602Zm6.8403 0c-.2199-.002-.4426.05-.6504.1602-.665.3526-.9181 1.1766-.5645 1.84.1669.313.6971-.1955 1.0079-.086.3476.1224.4867.8347.838.6485.6649-.3527.916-1.1767.5624-1.84-.243-.456-.7096-.7184-1.1934-.7227Zm-9.7565 1.418a.8768.8768 0 0 0-.877.877c0 .4846.3925.877.877.877a.8768.8768 0 0 0 .877-.877.8768.8768 0 0 0-.877-.877zm12.6434 0c-.4845 0-.879.3925-.879.877 0 .4846.3945.877.879.877a.8768.8768 0 0 0 .877-.877.8768.8768 0 0 0-.877-.877zM8.7927 11.459c-.179-.003-.2793.1107-.2793.416 0 .8097.3874 2.125 1.4279 2.924.207-.7123 1.3453-1.2832 1.5079-1.2012.2315.1167.2191.4417.6074.7266.3884-.285.374-.6098.6056-.7266.1627-.082 1.3009.4889 1.5079 1.2012 1.0404-.799 1.4278-2.1144 1.4278-2.924 0-1.2212-1.583.6402-3.5413.6485-1.4686-.0061-2.7266-1.0558-3.2639-1.0645zM4.312 14.4768c.5792.365 1.6964 2.2751 2.1056 3.0177.1371.2488.371.3536.582.3536.4188 0 .7465-.4138.0391-.9395-1.0636-.791-.6914-2.0846-.1836-2.1642a.4302.4302 0 0 1 .0664-.004c.4616 0 .666.7892.666.7892s.5959 1.4898 1.6213 2.508c.942.9356 1.062 1.703.4961 2.6661-.0164-.004-.0159.0236-.1484.2149-.1853.2673-.4322.4688-.7188.6152-.5062.2269-1.1397.2696-1.7833.2696-1.037 0-2.1017-.1824-2.6975-.336-.0293-.0075-3.6505-.9567-3.1916-1.8224.0771-.1454.2033-.2031.3633-.2031.6463 0 1.823.9551 2.3283.9551.113 0 .196-.0865.2285-.2031.2249-.8045-3.2787-1.0522-2.9846-2.1642.0519-.1967.193-.2757.3907-.2754.854 0 2.7704 1.4923 3.172 1.4923.0307 0 .0525-.0085.0645-.0274.2012-.3227.1096-.5865-1.3087-1.4395-1.4182-.8533-2.4315-1.329-1.8653-1.9416.0651-.0707.1574-.1015.2695-.1015.8611.0002 2.8948 1.84 2.8948 1.84s.5487.5683.8809.5683c.0762 0 .1416-.0315.1855-.1054.2355-.3946-2.1858-2.2183-2.3224-2.971-.0926-.51.0641-.7676.3555-.7676-.0006.008.1701-.0285.4942.1759zm16.2257.5918c-.1366.7526-2.5579 2.5764-2.3224 2.9709.044.074.1092.1055.1855.1055.3321 0 .881-.5684.881-.5684s2.0336-1.8397 2.8947-1.84c.1121 0 .2044.0308.2695.1016.5662.6125-.447 1.0882-1.8653 1.9415-1.4183.853-1.51 1.1168-1.3087 1.4396.012.0188.0337.0273.0644.0273.4016 0 2.3181-1.4923 3.1721-1.4923.1977-.0002.3388.0787.3907.2754.294 1.112-3.2095 1.3597-2.9846 2.1642.0325.1166.1156.2032.2285.2032.5054 0 1.682-.9552 2.3283-.9552.16 0 .2862.0577.3633.2032.459.8656-3.1623 1.8149-3.1916 1.8224-.5958.1535-1.6605.336-2.6975.336-.6351 0-1.261-.0409-1.7638-.2599-.2949-.1472-.5488-.3516-.7383-.625-.0411-.0682-.1026-.1476-.1426-.205-.5726-.9679-.455-1.7371.4903-2.676 1.0254-1.0182 1.6212-2.508 1.6212-2.508s.2044-.7891.666-.7891a.4318.4318 0 0 1 .0665.0039c.5078.0796.88 1.3732-.1836 2.1642-.7074.5257-.3797.9395.039.9395.211 0 .445-.1047.5821-.3535.4092-.7426 1.5264-2.6527 2.1056-3.0178.5588-.3524.99-.1816.8497.5918z};}}}
\author{%
  Nityanand Mathur\textsuperscript{*}\qquad
  Hamees Sayed\textsuperscript{**}\qquad
  Ayush Pratap Singh\\[11pt]
  \large\tensorviz\\[10pt]
  \normalfont\footnotesize
  \githubicon\hspace{0.25em}\href{https://github.com/nityanandmathur/aspect-d}{\texttt{nityanandmathur/aspect-d}}\quad
  \hficon\hspace{0.25em}\href{https://huggingface.co/nityanandmathur/aspect-d-masked-diffusion-tts}{\texttt{nityanandmathur/aspect-d-masked-diffusion-tts}}\\[-10pt]%
}

\begin{document}
\maketitle
{\renewcommand{\thefootnote}{*}\footnotetext{Work done outside of Blackstar Inc.\qquad\textsuperscript{**}Work done at Smallest AI.}}

\begin{abstract}
A diffusion language model buys sequential computation twice: depth, paid once in
parameters, and denoising steps, rented per utterance at serve time. We ask whether the
rented kind helps every capability equally. We train an iso-parameter width--depth grid of
\Nconfigs\ masked-diffusion codec TTS models (\Nseeds\ seeds, \NminN--\NmaxN\,M non-embedding
parameters) on \NtrainHours\,h of Emilia-EN and sweep the refinement budget $T$ at inference
only, evaluating cross-sentence zero-shot synthesis on \NevalItems\ items from
\NevalSpeakers\ held-out speakers. Our central result is model-free and referenced to floors
we \emph{measure} rather than to zero: the word error of an ASR system on the real
recordings, and the speaker similarity of a codec round trip. Against those floors,
refinement from $\tc{T}{=}1$ to $\tc{T}{=}16$ closes \NgapWER\% of the reachable
intelligibility range but only \NgapSIM\% of the reachable identity range, a
\NgapRatio\ asymmetry that holds between \NgapRatioLo\ and \NgapRatioHi\ under every
monotone reparameterisation of the error we tried. Retraining the same runs to $3\times$ and $6\times$ the
schedule attenuates that factor monotonically (\NgapThirtykRatio, \NgapNinetykRatio,
\NgapEightykRatio) while never reversing it, because refinement's intelligibility share
saturates (\NsatWER\,pp across the last doubling, and \NsatWERasrTwo\,pp under a
second recogniser from a different family) while its identity share keeps rising by
\NsatSIM\,pp. The two capabilities have different bottlenecks, and we therefore offer the
ordering, not the factor, as the result. We also report the pre-registered
power-law fit, and we report that the step-exponent contrast it yields
($\Delta\tau=\tau_{\mathrm{WER}}-\tau_{\mathrm{SIM}}=\Ndtau$) changes sign under those same
reparameterisations: it is a property of the error coordinate, and we therefore decline to
build the paper on it. The limit is on what \emph{refinement} buys, not on what inference
buys. At a matched number of function evaluations (NFE), best-of-$K$ search moves speaker similarity where refinement cannot,
and \NmencNfam\ speaker-encoder families outside the selector's lineage agree, with per-item
win rates of \NmencLo--\NmencHi\% and every interval above one half. Depth and steps are
\emph{not} interchangeable: the substitution form $B(d\,T^{\kappa})^{-\beta}$ fits worse than
the separable form by $\Delta\mathrm{AICc}=+\NaiccSSWER$. Finally, \NcodecShare\% of the
remaining identity gap belongs to the codec, not the model, so most of what identity has
left to gain is not the generator's to win.
\end{abstract}

\section{Introduction}
A diffusion language model can buy sequential computation twice. It \emph{owns} depth:
$d$ transformer blocks, paid once in parameters and then on every forward pass forever.
It \emph{rents} refinement steps: $T$ denoising passes over the same sequence, paid per
utterance and adjustable at serve time. An autoregressive model has only the first lever;
its serial cost per token is fixed by the architecture the moment training ends. If rented
serial compute substitutes for owned serial compute, a deployment can trade parameters for
steps after the fact. That trade is available only for the capabilities that steps improve.

Zero-shot text-to-speech is a clean probe: intelligibility is compositional (the right
phonetic content in the right order, read out by word error rate), while speaker identity is
representational (timbre carried from a short prompt, read out by SIM-o). The two need not
respond to the same knob.

We train one iso-parameter width--depth grid of masked-diffusion codec language models and
then sweep $T$ at inference only, which yields a full $(w, d, T)$ response surface per metric
at zero additional training cost. Four hypotheses were pre-registered before the first run:
that refinement is metric-selective (H-D2, primary), that steps and depth exchange at a rate
$\kappa$ (H-D3), that shape matters at fixed parameter count (H-D1), and that small budgets
predict large ones (H-D4). Table~\ref{tab:prereg} gives every verdict, including the failures.
Eq.~\eqref{eq:models} states the two fitted forms: one separable ($M_{\mathrm{sep}}$) in width $w$, depth $d$ and steps $T$,
one ($M_{\mathrm{sub}}$) in which steps substitute for depth through the product $d\,T^{\kappa}$.

\begin{equation}
\label{eq:models}
\begin{aligned}
\text{separable:}\quad &\mathrm{err}_m(w,d,T) = E_m + A_m w^{-\alpha_m} + B_m d^{-\beta_m} + C_m T^{-\tau_m}\\[-1pt]
\text{substitution:}\quad &\mathrm{err}_m(w,d,T) = E_m + A_m w^{-\alpha_m} + B_m \big(d\,T^{\kappa_m}\big)^{-\beta_m}
\end{aligned}
\end{equation}

\noindent Here $E_m$ is irreducible error, $A_m, B_m, C_m$ are amplitudes, $\alpha_m, \beta_m, \tau_m$ exponents and $\kappa_m$ the exchange rate. The pre-registered test is supported in the coordinate it declared:
$\Delta\tau = \Ndtau$, run-level bootstrap CI \Ndtauci, excluding zero. But $\tau$ is only
guaranteed invariant to \emph{affine} rescalings of the error, which is what our
pre-registration checked, and $\Delta\tau$ changes sign under monotone non-affine ones,
$\log(1+\mathrm{err})$ and $\sqrt{\mathrm{err}}$ among them (\S\ref{sec:scope}). A quantity
whose sign depends on whether one writes WER or $\sqrt{\mathrm{WER}}$ describes a
coordinate, not a system. We therefore restate the finding in a form that does not move:
against floors we measure, refinement closes \NgapWER\% of the reachable intelligibility
range and \NgapSIM\% of the reachable identity range. Depth and steps, meanwhile, do not
exchange at a rate one could design around: the substitution form fits \emph{worse} than the
separable one ($\Delta\mathrm{AICc}=+\NaiccSSWER$ for WER, $+\NaiccSSSIM$ for SIM-o), and
$\kappa$ runs to its bound, so no single ratio holds across the grid. Extra inference compute
is a targeted instrument, not a substitute for architecture.

\noindent\textbf{Contributions.}\hspace{0.4em}\cnum{1}~The first per-metric $(\wc{w},
\dc{d}, \tc{T})$ scaling surface for speech generation: \Nsurface\ points per metric from
\Nruns\ trained models, released with the training and analysis code.
\cnum{2}~A floor-referenced, model-free measurement of what refinement buys per capability
(\NgapWER\% versus \NgapSIM\% of the reachable range), together with the scope conditions
under which the usual exponent-based statement of the same effect fails: a
reparameterisation, a fitted range, and a goodness-of-fit test its own criterion rejects.
\cnum{3}~A per-axis account of what moves speaker identity, measured against a codec
ceiling: training compute and test-time selection move it where refinement does not, while
longer prompt context and rate matching buy nothing, with the selection result confirmed by
\NmencNfam\ encoder families outside the selector's lineage.

\section{Related work}
\textbf{Masked diffusion and TTS.}\hspace{0.4em}Masked diffusion language models
\citep{d3pm,sedd,mdlm} are trained at scale \citep{nie2025scaling,llada}, but their scaling
is reported against training compute.
The masked-generative recipes that made codec-token TTS practical
\citep{soundstorm,maskgit,maskgct,naturalspeech3} fix a step budget by engineering judgement
and ablate it at a few settings; none maps error against steps as a law.
In MaskGCT's step sweep, WER improves more than speaker similarity, a qualitative form of
our asymmetry.
Flow-matching TTS \citep{voicebox,e2tts,f5tts} also refines iteratively.
Autoregressive speech-token TTS \citep{valle,seedtts,cosyvoice} has been scaled in
parameters, data and verifier-guided search \citep{llasa}, which separates neither width
from depth nor either from steps.
Classical scaling work aggregates the same way \citep{kaplan,chinchilla}.

\textbf{Test-time compute.}\hspace{0.4em}Test-time scaling is well studied in reasoning
\citep{testtime}, where the payoff is a single correctness metric, and in image diffusion,
where search keeps paying after step gains flatten \citep{ma2025inference}.
A common form reranks sampled candidates with a verifier \citep{cobbe2021verifiers}.
In TTS, beam search on a speaker-similarity verifier raises SIM while WER can lag the no-search baseline \citep{llasa}.

\textbf{Depth and width.}\hspace{0.4em}At fixed size, loss depends only weakly on shape
\citep{kaplan}, yet depth helps more than width only up to a depth logarithmic in width
\citep{levine2020depth}.
Maximal-update parameterisation lets hyperparameters transfer across width \citep{mup}.
Weight-tied recurrence instead trades parameters for iterations of depth \citep{dehghani2019universal,geiping2025recurrent}.
We ask whether extra inference compute helps different capabilities \emph{differently}.
Over one width--depth grid, with steps swept at inference only, we measure the effect of
refinement and of best-of-$K$ search on intelligibility and on identity against measured
floors.

\begin{figure}[htbp]
  \centering
  \resizebox{0.86\linewidth}{!}{
\begin{tikzpicture}[
  font=\small,
  box/.style   = {rounded corners=2pt, draw=black!55, fill=white, inner sep=4pt,
                  align=center},
  cond/.style  = {box, fill=black!4, text width=32mm},
  core/.style  = {box, draw=dcol, line width=0.8pt, fill=dcol!6, text width=38mm},
  samp/.style  = {box, draw=tcol, line width=0.8pt, fill=tcol!6, text width=38mm},
  codec/.style = {box, fill=black!4, text width=15mm},
  ar/.style    = {-{Latex[length=1.7mm]}, draw=black!60, line width=0.55pt},
  vis/.style   = {fill=black!14},
  msk/.style   = {fill=tcol!35},
  cmt/.style   = {fill=wcol!30},
]

\node[cond] (txt) at (0,1.5)
     {prompt transcript $+$ target text\\[-1pt]
      \scriptsize eSpeak phonemes, always visible};
\node[cond] (aud) at (0,0)
     {3\,s prompt audio\\[-1pt]\scriptsize Mimi encode, 8 levels, always visible};
\node[cond, fill=white, draw=black!25] (len) at (0,-1.45)
     {\scriptsize target length $=$ chars $\cdot$ sec/char $\cdot\,12.5$ frames};

\begin{scope}[shift={(3.2,-0.8)}, x=2.3mm, y=1.9mm]
  \foreach \l in {0,...,7}{
    \foreach \f in {0,...,3}{ \draw[vis, draw=black!30] (\f,\l) rectangle ++(1,1); }
    \foreach \f in {4,...,11}{
      \ifnum\l<3 \draw[cmt, draw=black!30] (\f,\l) rectangle ++(1,1);
      \else      \draw[msk, draw=black!30] (\f,\l) rectangle ++(1,1);\fi }
  }
  \draw[decorate, decoration={brace, amplitude=3pt}] (0,8.4) -- (4,8.4)
        node[midway, above=1.5pt, font=\scriptsize] {prompt};
  \draw[decorate, decoration={brace, amplitude=3pt}] (4,8.4) -- (12,8.4)
        node[midway, above=1.5pt, font=\scriptsize] {generated};
\end{scope}
\node[font=\scriptsize, anchor=north, align=center] at (4.6,-1.15)
     {8 levels $\times$ frames\\[-2pt]
      \textcolor{tcol}{MASK} / \textcolor{wcol}{committed} / visible};

\node[core] (bb) at (8.6,0.05)
     {bidirectional masked-diffusion\\[-2pt] transformer\\[1pt]
      \textcolor{dcol}{depth $d$} $\times$ \textcolor{wcol}{width $w$},
      $N = 12\,d\,w^{2}$\\[1pt]
      \scriptsize RMSNorm, RoPE, no causal mask,\\[-3pt]\scriptsize no CFG};
\node[font=\scriptsize, align=center, text=dcol, anchor=north] at (8.6,-1.62)
     {\emph{owned} serial compute:\\[-2pt] paid once, in parameters};

\node[samp] (sm) at (13.3,0.05)
     {MaskGIT confidence decode\\[1pt]
      level $1 \to 8$, \textcolor{tcol}{$T$ steps per level}\\[1pt]
      \scriptsize cosine schedule, Gumbel noise\\[-3pt]\scriptsize $1\!\to\!0$, NFE $= 8T$};
\node[font=\scriptsize, align=center, text=tcol, anchor=north] at (13.3,-1.62)
     {\emph{rented} serial compute:\\[-2pt] paid per utterance, set at serve time};

\node[codec] (dec) at (17.1,0.05) {Mimi\\decoder};
\node[font=\scriptsize, anchor=north] (wav) at (17.1,-0.85) {waveform};

\draw[ar] (txt.east) -- ++(0.3,0) |- (2.75,0.6);
\draw[ar] (aud.east) -- ++(0.3,0) |- (2.75,0.0);
\draw[ar] (len.east) -- ++(0.3,0) |- (2.75,-0.6);
\draw[ar] (6.35,0.05) -- (bb.west);
\draw[ar] (bb.east)  -- (sm.west);
\draw[ar] (sm.east)  -- (dec.west);
\draw[ar] (dec.south) -- (wav.north);
\draw[ar, dashed, draw=tcol] (sm.north) -- ++(0,0.55) -| (bb.north)
      node[pos=0.28, above, font=\scriptsize, text=tcol]
      {re-mask the lowest-confidence cells and repeat};
\end{tikzpicture}}
  \caption{\emph{The two ways a masked-diffusion TTS model buys sequential computation.}
  Phonemes \citep{phonemizer} and a 3-second prompt are always visible; the generated region of the
  $8 \times \text{frames}$ token grid starts fully \tc{masked}. The backbone spends
  \dc{depth} (\emph{owned} serial compute, fixed at training time), and the sampler
  spends \tc{$T$ refinement steps per codebook level} (\emph{rented} serial compute,
  chosen at serve time). One training grid over $(\wc{w}, \dc{d})$ therefore yields a whole
  $(\wc{w}, \dc{d}, \tc{T})$ surface at inference. This paper asks whether the rented kind
  helps every capability equally.}
  \label{fig:pipeline}
\end{figure}
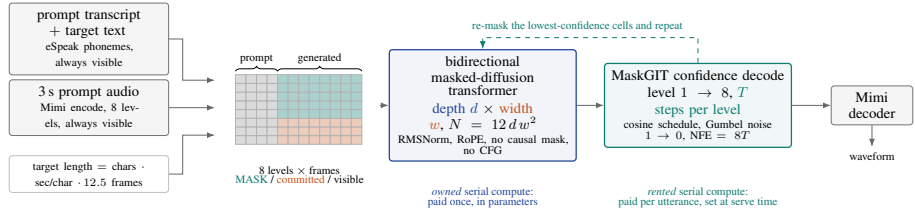

\section{Method}
\paragraph{Backbone and objective.}
The backbone is a bidirectional (non-causal) pre-LN transformer over Mimi \citep{moshi}
codec tokens at 12.5\,Hz, using the first 8 residual codebooks (2048 entries each, plus MASK
and PAD). It is parameterised by \wc{width $w$} and \dc{depth $d$}, with head dimension 64,
GELU MLP ratio 4, RMSNorm, RoPE \citep{rmsnorm,rope} and no weight tying, so the non-embedding parameter count is
$N = 12\,d\,w^{2}$. One embedding table per codebook is summed at the input and one linear
head per codebook reads out. Training follows SoundStorm-style \citep{soundstorm}
coarse-to-fine masking: per example we draw a level $l \sim \mathcal{U}\{1..8\}$ and a ratio
$t \sim \mathcal{U}(0,1]$; levels below $l$ stay visible, level $l$ is masked i.i.d.\ with
probability $t$, and levels above $l$ are entirely MASK. The loss is cross-entropy on the
masked cells of level $l$ only. A phoneme prefix and a 3-second prompt of codec tokens are
always visible. The model has no speaker encoder, so the backbone is the only identity
pathway, and it is trained without classifier-free guidance (CFG; \citealp{cfg}), which would otherwise confound
the step axis. The target length is fixed before generation from the text length and a corpus
character rate ($\Nsecperchar$\,s/char).

\paragraph{Data and task.}
All models are trained on a \NtrainHours\,h speaker-stratified subset of Emilia-EN
\citep{emilia} covering more than \NtrainSpk\ speakers, with \NheldoutSpk\ held out entirely.
Evaluation is cross-sentence zero-shot synthesis (prompt and target are \emph{different}
utterances from the same held-out speaker, so copying the prompt's content does not score)
on \NevalItems\ items from \NevalSpeakers\ unseen speakers, targets 4--15\,s. Two measured
floors anchor every number we report: the word error Whisper-large-v3 \citep{whisper} makes on the real
recordings of those items (\NwerFloor), and the speaker similarity of a Mimi round trip (encode, then decode) of
the real target with no model in the loop (\NsimCeiling, the \emph{codec ceiling}). Neither is reachable by a better
generator, and referencing gains to zero instead of to these floors is what makes the two
metrics look incomparable.

\paragraph{Frozen sampler.}
Inference is MaskGIT-style confidence decoding \citep{maskgit}, run level by level with
exactly \tc{$T$ refinement steps per level}, so the number of function evaluations is NFE~$= 8T$. A cosine schedule sets how many
cells remain masked after each step, tokens are drawn at temperature 1.0, and confidences are
log-probabilities plus Gumbel noise annealed $1 \to 0$; committed cells stay committed. The
sampler is identical for every config, seed and $T$, and generation noise is keyed so that
$T$ is the only difference between surface points of one run.

\paragraph{Design and statistics.}
The grid holds $\Nconfigs$ configs: three iso-$N$ budgets $\times$ five aspect ratios, each
trained with $\Nseeds$ seeds for 30{,}000 steps at effective batch 256 with maximal-update parameterisation ($\mu$P) \citep{mup}
(base width 256, base LR $\Nbaselr$, transfer checked by five-point LR sweeps). We sweep
$T \in \{\NTgrid\}$ at evaluation only, which gives $\Nsurface$ surface points per metric from
$\Nruns$ trained models. The metric stack is WER (Whisper-large-v3 \citep{whisper}, greedy),
SIM-o (WavLM-large \citep{wavlm} cosine against the \emph{original} prompt waveform, following
\citet{voicebox,seedtts}), a frozen degenerate-output rule (transcript empty, under 30\% of the reference length, or a 4-gram repeated $6\times$), and UTMOS \citep{utmos} for predicted naturalness. The error
transforms are $\mathrm{err}_{\mathrm{WER}} = \mathrm{WER}$ and
$\mathrm{err}_{\mathrm{SIM}} = 1 - \text{SIM-o}$, computed over all items including
degenerates. We fit both forms of Eq.~\eqref{eq:models} by weighted non-linear least squares
with 32 multi-starts. The intervals on $\Delta\tau = \tau_{\mathrm{WER}} -
\tau_{\mathrm{SIM}}$, $\kappa$ and $\Delta\rho = \alpha_{\mathrm{SIM}}/\beta_{\mathrm{SIM}} -
\alpha_{\mathrm{WER}}/\beta_{\mathrm{WER}}$ are 95\% percentile intervals from a bootstrap \citep{efron1979bootstrap} of
$\Nbootreps$ replicates that resamples at the \emph{run} level and recomputes every $T$ point.
The run level is the right unit because the five $T$ points of one run share a trained model,
so resampling surface points would understate their correlation.

\section{Results}
\subsection{What refinement buys, measured against floors}
\label{sec:gapclosed}
The comparison that matters is not how much error each metric sheds (WER and $1-$SIM-o
have no common unit) but how much of the error that is \emph{reachable at all} each one
sheds. Table~\ref{tab:gapclosed} reports that fraction directly from the measurements, with
no fitted model. Going from one-step decoding to the deployable $\tc{T}{=}16$ default,
refinement closes \NgapWER\% (CI \NgapWERci) of the distance from $\tc{T}{=}1$ to the automatic speech recognition (ASR)
floor, and \NgapSIM\% (CI \NgapSIMci) of the distance to the codec ceiling: a ratio of
\NgapRatio\ (CI \NgapRatioci). Refinement very nearly exhausts what is available to it on
intelligibility while leaving most of the identity range untouched (Fig.~\ref{fig:main}).

Two properties make this the statement we build on. First, it is stable in $\tc{T}$: we
extended every run to $\tc{T}{=}\NgapTmax$, and over that $16\times$ range the ratio stays between \NgapRatioTLo\ and \NgapRatioTHi\ while $\Delta\tau$ halves
(\S\ref{sec:scope}). Quadrupling the budget past the deployable default takes
intelligibility from \NgapWER\% to \NgapWERext\% but identity only from \NgapSIM\% to
\NgapSIMext\% (CI \NgapSIMextci), and pushing the largest budget to $\tc{T}{=}128$ moves WER
by \NTmaxWERdelta\ and SIM-o by \NTmaxSIMdelta: the refinement axis has stopped paying and
begun, slightly, to cost. Second, it is stable in the coordinate: because it is a ratio of differences
along one axis, a monotone reparameterisation moves numerator and denominator together, so
across $\log(1+\mathrm{err})$, $\sqrt{\mathrm{err}}$ and $\mathrm{err}^{2}$ the ratio stays
between \NgapRatioLo\ and \NgapRatioHi\ where $\Delta\tau$ changes sign.

\subsection{The two capabilities have different bottlenecks}
\label{sec:trend}
The asymmetry is not an intrinsic property of the two capabilities, and saying so is the
paper's main claim rather than a caveat on it. We retrained the same \NtrendRuns\ runs
(configs C1/C3/C5, i.e.\ the shallowest, middle and deepest \Nbudgetc\ shapes, $\times$ \Nseeds\ seeds, identical at every budget) for $3\times$ and
$6\times$ the original schedule and repeated the measurement (Table~\ref{tab:trend}). The
ratio attenuates monotonically (\NgapThirtykRatio, \NgapNinetykRatio, \NgapEightykRatio)
with non-overlapping intervals, and stays strictly above one throughout.

What drives it is visible in the two shares separately. \emph{Refinement's intelligibility
share saturates and stays saturated}: across the $2\times$ step from 90k to 180k it moves by
\NsatWER\ percentage points, and a second recogniser from a different family
(\NasrTwoName\ \citep{wav2vec2,xu2021selftraining}, a connectionist temporal classification (CTC) model) puts the same figure at \NsatWERasrTwo\ points:
two systems that disagree by a factor of $2.4$ on absolute word error agree to a tenth of a
point that further training buys no intelligibility. At 180k our models reach that
recogniser's word error on the \emph{real recordings} to within \NasrTwoFloor\ absolute, so
there is nothing left to close. \emph{The identity share, over the same step, rises by
\NsatSIM\ points and is still rising.}

The selectivity therefore measures how far from converged each capability is, not what
refinement is able to buy. Intelligibility is bounded by refinement and reaches that bound
early; identity is bounded by training compute and by the codec, and refinement cannot reach
either. This is why the ordering, and not the factor, is what we report: the factor is a
property of the training budget, and under-training inflates it. Three budgets do not
determine whether the ratio converges above one or to one, and we do not extrapolate.

The grid resolves shape differences well above seed noise: within each budget the spread of
config means at $\tc{T}{=}16$ is at least twice the pooled seed standard deviation, so the
pre-registered power criterion \Ngfour\ in all three budgets.

\subsection{The same effect as a fitted exponent, and why we do not lead with it}
\label{sec:scope}
Our pre-registration declared a power-law surface (Eq.~\ref{eq:models}) and a primary test on
its step exponents. That test passes: $\tau_{\mathrm{WER}} = \Ntauwer$ (CI \Ntauwerci)
against $\tau_{\mathrm{SIM}} = \Ntausim$ (CI \Ntausimci), giving $\Delta\tau = \Ndtau$, CI
\Ndtauci, excluding zero (H-D2). Three scope conditions bound what it means, and we state
them because each one weakens a claim we would otherwise have made.

\cnum{1}~\emph{Coordinate.} $\tau$ is invariant to affine rescalings of the error, which is
what we verified before running; it is not invariant to monotone non-affine ones. Refitting
the identical measurements under $\log(1+\mathrm{err})$, $\sqrt{\mathrm{err}}$ and a
cap at $\mathrm{err}=1$ sends $\Delta\tau$ \emph{negative} in \NdtauNflip\ of the
reparameterisations we tried (Table~\ref{tab:scope}). Its sign is a fact about the coordinate.
\cnum{2}~\emph{Range.} $\tau$ summarises curvature over the window it is fitted on, and is
not an identified asymptotic exponent. If we extend every run to $\tc{T}{=}\NgapTmax$ and
refit on nested windows, $\Delta\tau$ falls from \NdtauRangeEight\ over $\tc{T}\le8$ to
\NdtauRangeSixteen\ over the declared $\tc{T}\le16$ and \NdtauRangeSixtyfour\ over
$\tc{T}\le64$: it roughly halves each time the window doubles. The declared value is a local
slope at the edge of the declared grid. \cnum{3}~\emph{Fit.} Judged by the weighted
criterion the fit itself minimises, the declared form is rejected as a generative model:
$\chi^{2}/\mathrm{dof} = \NchiWER$ for WER and \NchiSIM\ for SIM-o. We use it as a compact
description of curvature and read no mechanism from its parameters. The declared interval
also holds the $1/\mathrm{SE}^2$ weights (SE: standard error over seeds) fixed at their observed-data values; resampling the
weights as well widens it to \NdtauHonestci.

None of this touches \S\ref{sec:gapclosed}, which is model-free. It is why
\S\ref{sec:gapclosed} is the result we lead with.

\begin{table}[htbp]
  \centering
  \caption{\emph{What refinement buys, as a fraction of what is reachable.} Share of the distance from one-step decoding to a measured reference, Whisper's word error on the real recordings (\NwerFloor) and the speaker similarity of our codec's round trip (\NsimCeiling), closed by step budget $\tc{T}$, on the full \Nruns-run grid. Percentile CIs from a \Nbootreps-replicate run-level bootstrap. No model is fitted, and the ratio is invariant to monotone reparameterisation of the error (\NgapRatioLo--\NgapRatioHi, Table~\ref{tab:scope}).}
  \label{tab:gapclosed}
  \footnotesize
\begin{tabular}{lccc}
  \toprule
  & \multicolumn{2}{c}{fraction of reachable range closed} & \\
  \cmidrule(lr){2-3}
  $\tc{T}$ & intelligibility (WER) & identity (SIM-o) & ratio \\
  \midrule
    2 & 32.2\% [29.7, 34.4] & 20.0\% [19.1, 20.9] & 1.61$\times$ \\
    4 & 69.8\% [66.0, 73.1] & 35.7\% [34.3, 37.0] & 1.96$\times$ \\
    8 & 81.8\% [78.3, 84.8] & 43.5\% [42.1, 44.9] & 1.88$\times$ \\
    16 & 86.2\% [82.8, 89.1] & 46.4\% [45.0, 47.8] & 1.86$\times$ \\
    32 & 87.9\% [84.9, 90.5] & 47.5\% [46.2, 48.8] & 1.85$\times$ \\
    64 & 88.4\% [85.3, 91.0] & 47.9\% [46.6, 49.2] & 1.85$\times$ \\
  \bottomrule
\end{tabular}

\end{table}

\begin{table}[htbp]
  \centering
  \caption{\emph{The exponent contrast is a property of the error coordinate.} The same
  \Nsurface\ measurements, refitted after a monotone reparameterisation of both errors. $\tau$
  is invariant to the affine map, as our original (v1.0) pre-registration assumed; under the non-affine maps
  $\Delta\tau$ changes sign, while the model-free ratio of Table~\ref{tab:gapclosed} does not.
  This is why the paper's claim is stated as a share of a reachable range rather than as an
  exponent difference.}
  \label{tab:scope}
  \footnotesize
\begin{tabular}{lcccc}
  \toprule
  error coordinate & $\tau_{\mathrm{WER}}$ & $\tau_{\mathrm{SIM}}$ & $\Delta\tau$ & gap-closed ratio \\
  \midrule
    identity \emph{(pre-registered)} & 0.8362 & 0.7260 & +0.1102 & 1.86$\times$ \\
    log(1+err) & 0.6665 & 0.6913 & -0.0248 & 1.85$\times$ \\
    sqrt(err) & 0.4325 & 0.6842 & -0.2517 & 1.68$\times$ \\
    err capped at 1 & 0.7093 & 0.7260 & -0.0167 & n/a \\
    err squared & 1.5613 & 0.8122 & +0.7491 & 1.82$\times$ \\
    affine 3+2*err (v1.0 check) & 0.7600 & 0.7249 & +0.0351 & n/a \\
  \bottomrule
\end{tabular}

\end{table}

\begin{table}[htbp]
  \centering\footnotesize
  \caption{\emph{Refinement's intelligibility share saturates; its identity share does not.}
  The same \NtrendRuns\ runs (C1/C3/C5 $\times$ \Nseeds\ seeds) retrained to three budgets and
  measured identically: share of the reachable range closed from $\tc{T}{=}1$ to
  $\tc{T}{=}16$. The second intelligibility column re-transcribes the identical audio with
  \NasrTwoName, a CTC recogniser sharing neither architecture, objective nor training corpus
  with Whisper; it is reported so that saturation cannot be an artefact of one recogniser.
  Intervals cluster the bootstrap on \emph{configuration} (seeds within a config are not
  draws from config space, so the effective $n$ is \NtrendBudgets) and resample evaluation
  items as a second dimension.}
  \label{tab:trend}
\begin{tabular}{lccccc}
  \toprule
  & \multicolumn{2}{c}{intelligibility share} & identity & & \\
  \cmidrule(lr){2-3}
  training & Whisper & \NasrTwoName & share & ratio & 95\% CI \\
  \midrule
    30k & 90.6\% & 88.5\% & 49.3\% & 1.84$\times$ & 1.73, 1.97 \\
    90k & 97.1\% & 99.2\% & 71.5\% & 1.36$\times$ & 1.28, 1.44 \\
    180k & 97.0\% & 99.1\% & 78.8\% & 1.23$\times$ & 1.18, 1.29 \\
  \bottomrule
\end{tabular}

\end{table}

\subsection{Robustness of the asymmetry}
The selectivity is not an artefact of scale, training compute or the metric stack
(Table~\ref{tab:robust}). A fourth parameter budget (\NmaxNfour\,M, ten further runs)
gives $\Delta\tau = \Ndtaufour$, CI \Ndtaufourci; retraining three configurations
$3\times$ longer gives \Ndtaunineok, CI \Ndtaunineokci; a different ASR gives \Ndtauasr\
and unbounded log-amplitudes \Ndtaulogamp. Extending $T$ to 64 leaves WER still falling
past the grid edge (\NwerTsixteenExt\ $\to$ \NwerTsixtyfour), so the gain measured at
$\tc{T}{=}16$ is a lower bound.

\begin{figure}[htbp]
  \centering
  \begin{minipage}{0.43\linewidth}\centering
    \includegraphics[width=\linewidth]{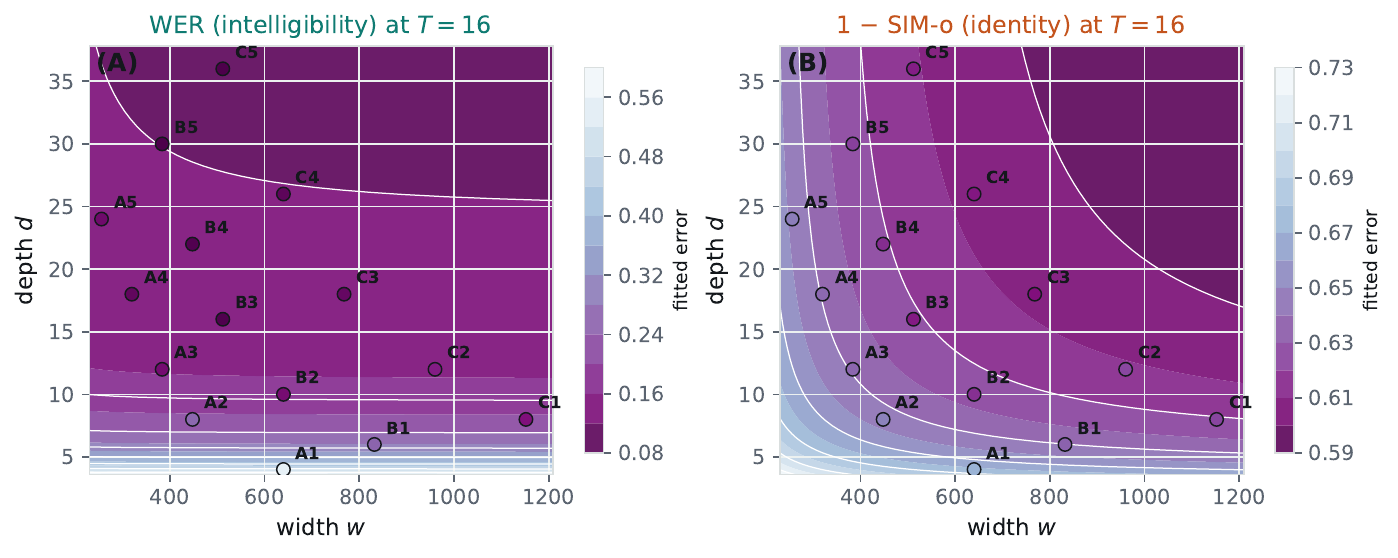}
  \end{minipage}\hfill
  \begin{minipage}{0.43\linewidth}\centering
    \includegraphics[width=\linewidth]{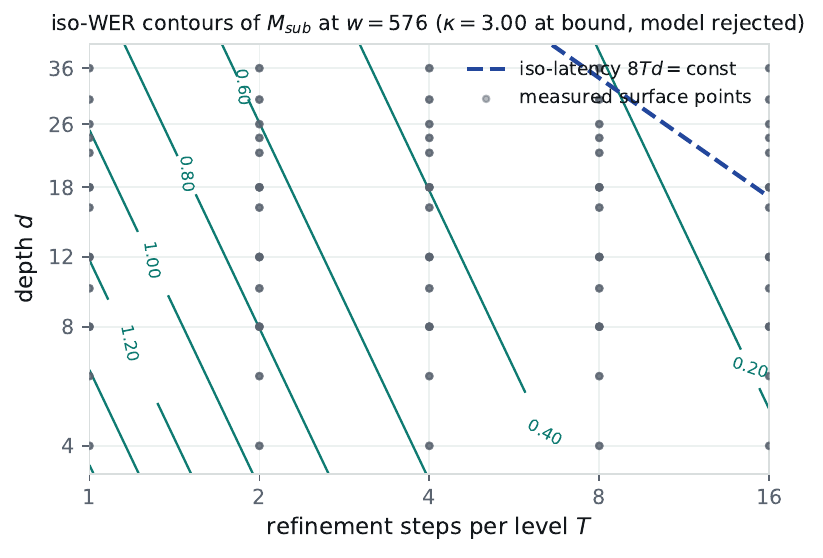}
  \end{minipage}
  \caption{\emph{Left:} iso-error contours in $(\wc{w}, \dc{d})$ at $\tc{T}{=}16$. The two
  capabilities do not share a preferred shape, but the width amplitude saturates its bound,
  so the anisotropy ratio $\rho$ is not identified. \emph{Right:} iso-WER contours in
  $(\log \tc{T}, \log \dc{d})$: a single substitution factor $d\,T^{\kappa}$ cannot reproduce
  both the depth ordering at fixed $T$ and the $T$ ordering at fixed depth.}
  \label{fig:diag}
\end{figure}

\paragraph{Depth and steps do not substitute.}
The substitution form of Eq.~\eqref{eq:models} fits the surface worse than the separable
form by the corrected Akaike information criterion: $\Delta\mathrm{AICc}(M_{\mathrm{sub}}-M_{\mathrm{sep}}) = +\NaiccSSWER$ for WER and
$+\NaiccSSSIM$ for SIM-o, where the pre-registered claim (H-D3) required $\le +4$, and
$\kappa$ sits at its bound in every fit. The honest reading is narrower than ``no exchange
rate exists''. A global factor $d\,T^{\kappa}$ is a strong constraint: it demands that one
ratio hold at every depth and every step count, and it is that constraint the data reject (Fig.~\ref{fig:diag}, right).
The separable fit still implies a \emph{local} rate, $\kappa \approx \tau/\beta =
\NkappaImpliedWER$ for WER, valid near the operating point and nowhere claimed to extend
beyond it. What does not survive is the design rule: one cannot trade depth for steps at a
fixed rate across the grid.

\paragraph{Shape matters, but the anisotropy ratio is not identified.}
At $\tc{T}{=}16$ the two-axis model beats the $N$-only model for both metrics
($\Delta\mathrm{AICc} = \NaiccFNWER$ / $\NaiccFNSIM$ against a $-4$ threshold; \citealp{aicc}), so error at
fixed parameter count depends on how those parameters are shaped (Fig.~\ref{fig:diag}, left). The direction of that
dependence does not resolve. The width amplitude saturates its pre-registered upper bound
in every fit, so only the product $A\,w^{-\alpha}$ is identified and $\alpha$ is not, which
leaves the anisotropy ratio $\rho = \alpha/\beta$ undetermined and the secondary
pre-registered claim (H-D1) not supported.

\dc{Depth} is identified, and it separates the capabilities sharply
($\beta_{\mathrm{WER}} = \NbetaBWER$ versus $\beta_{\mathrm{SIM}} = \NbetaBSIM$; Table~\ref{tab:exponents}): moving
from the shallowest shape to the best depth is worth \NdepthgainA, \NdepthgainB\ and
\NdepthgainC\ WER points across the three budgets, while SIM-o varies by under
\NsimspreadTsixteen. The depth optimum is interior only in the smallest budget; in the two
larger ones the argmin sits at the deepest shape tested, so those are lower bounds, not
interior optima.

\paragraph{Extrapolation and degenerate outputs.}
Fits on the two smaller budgets do not predict the largest (H-D4 refuted), and the frozen degenerate rule fires on \Ndegentsixteen\% of items at $\tc{T}{=}16$, concentrated at the shallowest shapes. Both are reported in full in the appendix (Fig.~\ref{fig:extrap}, Appendix~\ref{app:grid}).

\paragraph{Serving corollary.}
Per-layer latency is flat in width: measured at batch 1, $c_{\mathrm{layer}}$ is
\Nclayerrange\,ms for $w \in [256, 1152]$, because the device is launch-bound at this
scale. Serial latency is therefore $8\,T\,d\,c_{\mathrm{layer}}$, so at fixed parameter
count width is free and a latency budget constrains only the product $T\,d$. How to divide
that product between depth and steps is not determined here. The pre-registered rule, buy
depth before steps, is refuted: it holds at 5.9\% of the 40 latency budgets we test,
against a required 80\%. The reverse ordering is a boundary effect of a grid censored at
$T{=}16$, not a finding.

\begin{table}[htbp]
  \centering\footnotesize
  \caption{\emph{What buys speaker identity.} Absolute SIM-o gain per axis, set against the
  \NheadroomSim\ of headroom between our best system and a codec round trip of real audio.
  The headroom is a WavLM-large quantity, so every row is reported under that same encoder
  (final column); the cross-encoder agreement evidence for the search row is separated out
  into Table~\ref{tab:menc} rather than mixed into this column. Training compute and
  test-time selection buy the most. Guidance buys a little and charges for it in WER; rate
  matching and longer prompts buy nothing.}
  \begin{tabular}{llcll}
    \toprule
    axis & change & $\Delta$ SIM-o & cost & scorer \\
    \midrule
    training compute & 30k $\to$ 90k steps & \NtrainGainSim & $3\times$ training & WavLM-L \\
    \quad(further)    & 90k $\to$ 180k      & \NtrainDouble  & $2\times$ again & WavLM-L \\
    test-time search  & best-of-8 vs $\tc{T}{=}16$ & \NsearchVsDefault & \NsearchWERcost\ WER, $8\times$ NFE & WavLM-L \\
    parameters        & 20\,M $\to$ 276\,M  & \NparamGainSim & $14\times$ parameters & WavLM-L \\
    guidance          & $\gamma = 0.5$      & \NguideGainSim & \NguideWER\ WER points & WavLM-L \\
    rate matching     & per-item duration   & \NrateGainSim  & free & WavLM-L \\
    prompt context    & 3\,s $\to$ 9\,s     & none           & $+$\NctxWerCost\ WER points & WavLM-L \\
    \bottomrule
  \end{tabular}
  \label{tab:ledger}
\end{table}

\begin{figure}[htbp]
  \centering
  \includegraphics[width=0.86\linewidth]{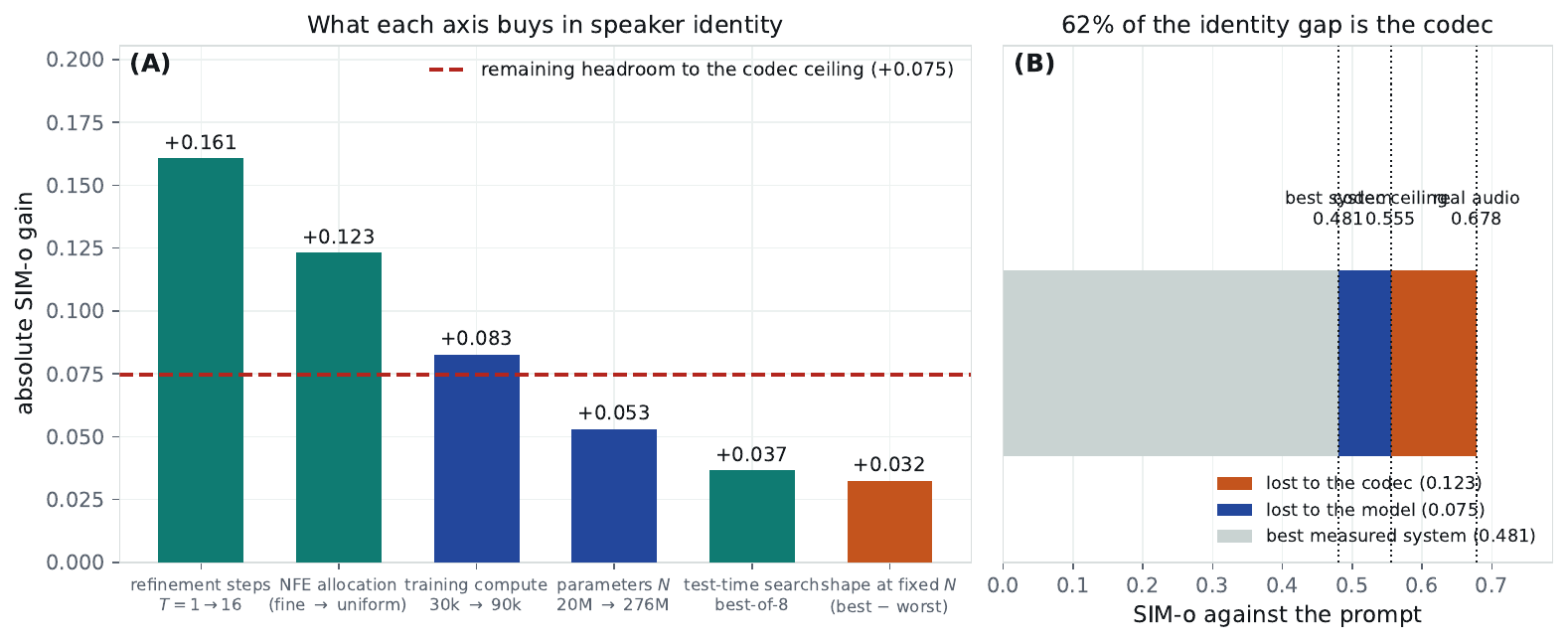}
  \caption{\emph{Identity is mostly not the model's to win.} \textbf{(A)}~what each axis buys
  in speaker similarity, against the headroom that remains to the codec ceiling; bars are
  coloured by which lever they are (\tc{rented at serve time}, \dc{owned in training},
  \wc{shape}). Refinement's bar is the largest only because it is measured from one-step
  decoding, which is not an operating point anyone ships; from the deployable
  $\tc{T}{=}16$ default it buys nothing further, and test-time search is what moves identity
  from there. \textbf{(B)}~of the distance between real audio and our best system,
  \NcodecShare\% is lost in the codec itself.}
  \label{fig:ledger}
\end{figure}

\paragraph{Training compute is the largest identity lever, and has not saturated.}
Retraining three configurations to 90k steps raises SIM-o by \NtrainGainSim\ (CI
\NtrainGainCI, \NtrainGainRuns\ runs, all positive), and a further doubling to 180k adds
\NtrainDouble\ (CI \NtrainDoubleCI) at a WER cost inside the $\pm2.0$-point guardrail.
Across $\{30\text{k}, 90\text{k}, 180\text{k}\}$ steps on one configuration, SIM-o runs
\NsimThirtyk\ $\to$ \NsimNinetyk\ $\to$ \NsimOneeightyk. Extrapolating the first interval
log-linearly predicts \NtrainPred\ at 180k against \NsimOneeightyk\ observed, a residual of
\NtrainResid: the axis is decelerating but still paying, and per unit it buys more identity
than any inference-time intervention we tested.

\paragraph{What buys identity, measured against a ceiling.}
If refinement is an intelligibility knob, what moves identity? We put every candidate lever
on one scale (Table~\ref{tab:ledger}, Fig.~\ref{fig:ledger}) and read it against a fixed bound. A codec round trip of real audio
scores \NsimRoundtrip, leaving \NheadroomSim\ of headroom above our best system.

Training compute is the largest lever: 30k${\to}$90k steps gains \NtrainGainSim\ (CI
\NtrainGainCI\ over \NtrainGainRuns\ runs), and a further doubling adds \NtrainDouble.
Test-time search is the second lever: best-of-$K$ sampling at matched NFE ($8T$ per
candidate), selecting with one speaker encoder and scoring with a second. It gains
\NsearchDelta\ over refinement at $\tc{T}{=}64$ (CI \NsearchCI, on \NsearchWin\% of items)
and \NsearchVsDefault\ over the deployable $\tc{T}{=}16$ default, at a cost of
\NsearchWERcost\ WER. Refinement remains the better spend for intelligibility: WER is
\NrefineWER\ under refinement against \NsearchWER\ under search. The gain comes from
selection, not from sampling: an arbitrary candidate scores \NrandomPick\ regardless of $K$.
It persists at \NsearchMaxN\,M non-embedding parameters (\NsearchD) and, on three runs, at
$3\times$ training compute (\NsearchNinetyk).

The remaining levers are small or empty. A $14\times$ parameter increase gains
\NparamGainSim. Training-free speaker-contrastive guidance (logits pushed away from those under a wrong speaker's prompt, with weight $\gamma$) gains \NguideGainSim\ and costs
\NguideWER\ WER points, so it trades one capability for the other (Table~\ref{tab:negative}). Rate-matched length
conditioning gains \NrateGainSim, and longer prompts gain nothing (Table~\ref{tab:context}): the best prompt is the
3\,s default seen in training, and beyond it the output stays fluent and speaker-consistent
while decoupling from the text. Identity is bought with training and with selection, not
with more refinement.

\emph{Where} the passes are spent matters as much as how many: at matched
NFE${}=32$, moving them to the coarsest codebook level changes WER from
\NnfeFineWER\ to \NnfeCoarseWER, at a cost in identity and naturalness
(Appendix~\ref{app:alloc}).

\begin{table}[htbp]
  \centering\footnotesize
  \caption{\emph{Refinement and search buy different capabilities at the same cost.} Matched
  NFE ($8T$ per candidate); best-of-$K$ candidates are selected by WavLM-SV and scored by a
  separate ECAPA encoder. Refinement wins intelligibility at every tier, and search wins
  identity at every tier.}
  \begin{tabular}{lccccc}
    \toprule
    NFE & refinement & search & $\Delta$ ECAPA & 95\% CI & WER (refine / search) \\
    \midrule
    128 & $\tc{T}{=}\NsrchLoT$ & $K{=}\NsrchLoK$ & \NsrchLod & \NsrchLoci & \NsrchLorw\ / \NsrchLosw \\
    256 & $\tc{T}{=}\NsrchMidT$ & $K{=}\NsrchMidK$ & \NsrchMidd & \NsrchMidci & \NsrchMidrw\ / \NsrchMidsw \\
    512 & $\tc{T}{=}\NsrchHiT$ & $K{=}\NsrchHiK$ & \NsrchHid & \NsrchHici & \NsrchHirw\ / \NsrchHisw \\
    \midrule
    \multicolumn{6}{l}{\footnotesize replicated at NFE 512 outside the grid:
    \NmaxNfour\,M budget \NscopeDd\ \NscopeDci; $3\times$ training \NscopeNd\ \NscopeNci} \\
    \bottomrule
  \end{tabular}
  \label{tab:search}
\end{table}

\begin{table}[htbp]
  \centering\footnotesize
  \caption{\emph{The search result is not an artefact of the selector's lineage.} The same
  \NmencItems\ items per run, \NmencRuns\ runs, rescored end-to-end by five speaker encoders
  spanning four model families. The selector is WavLM-large SV, so its own row is circular
  and is shown only for reference. A \emph{per-item win rate} (how often an encoder ranks
  the search output above the refinement output at matched NFE) is invariant to any
  monotone rescaling of that encoder's cosine, which a difference of means is not; we
  therefore compare win rates across encoders and report each raw $\Delta$ only beside its own
  scale. Every family, including the three whose absolute cosines fail our ground-truth gate (median same-speaker cosine $\ge 0.50$, cross-speaker $\le 0.25$)
  \emph{purely on scale} while separating speakers at AUC $\ge \NmencMinAuc$, puts search
  ahead with an interval clear of one half.}
\begin{tabular}{llccc}
  \toprule
  encoder & family / role & win rate & 95\% CI & raw $\Delta$ \\
  \midrule
    ECAPA-TDNN \citep{ecapa} & ECAPA, gated & 72.3\% & [70.6, 73.8] & +0.0444 \\
    x-vector \citep{xvector} & TDNN & 79.0\% & [77.5, 80.5] & +0.0075 \\
    GE2E d-vector \citep{ge2e} & LSTM d-vector & 69.1\% & [67.5, 70.8] & +0.0164 \\
    WavLM-base+ SV & WavLM, other scale & 64.6\% & [62.9, 66.3] & +0.0173 \\
    WavLM-large SV$^{\ast}$ & WavLM / \emph{selector} & 84.9\% & [83.6, 86.1] & +0.0803 \\
    \midrule
    \multicolumn{5}{l}{\footnotesize $^{\ast}$selects the candidates, so this row is circular and is not counted as agreement.} \\
  \bottomrule
\end{tabular}

  \label{tab:menc}
\end{table}

\section{Limitations and outlook}
Every model in the grid is under-trained. We cap non-embedding parameters at \NmaxN\,M and
training at 30{,}000 steps, so WER's saturation step $T^{*}_{\mathrm{WER}}$ is a lower bound. Part of the identity
gap is also not ours to close: a codec round trip of the target scores \NsimRoundtrip\
against \NsimRealAudio\ for the original waveform, so \NcodecShare\% of the distance to real
speech is lost in the codec before the model generates anything.

Our selector and our original scorer share a lineage (both are ECAPA-TDNN-family models,
correlating per item at $r$ in the range \NselCorrRange), so the search result was open to
the charge that it measured a shared representational bias rather than identity.
Table~\ref{tab:menc} closes that charge but does not make the measurement lineage-free: the
candidates are still \emph{chosen} by one encoder, and what the other four establish is that
the chosen candidates are better by their lights too, not that a different selector would
choose the same ones. Compute matching (Table~\ref{tab:search}) also covers the diffusion transformer only, so the
selector and scorer forward passes are not charged to the search arm.

The study covers one language, one codec and one corpus, with no classifier-free guidance. Target
length comes from a corpus median character rate rather than a per-item prediction. Two
fitted quantities are not identified: $\kappa$ sits on its bound in every fit, and the width
amplitude saturates its bound, which leaves $\Delta\rho$ undetermined. The $\Delta\tau$
interval reflects seed variation only, and widens to \NdtauHonestci\ once the fit's weights
are resampled along with the runs.

Table~\ref{tab:prereg} in Appendix~\ref{app:grid} gives every pre-registered verdict. Four
extension (H-E) and speaker-identity (H-S) hypotheses were refuted outright (H-E1b, H-E2, H-E4b, H-S4), three were not supported (H-D1,
H-D3, H-D4) and two were discordant between their pre-specified lenses (H-S1, H-S3); we name
them because a paper that quotes only the passing half of a split hypothesis cannot be
audited. Two censorings remain, both removable by a larger grid: the depth optimum sits at
the deepest shape tested in the two larger budgets, and WER is still falling at the top of
the fitted $T$ range.

\bibliographystyle{plainnat}
\bibliography{refs}

\appendix
\section{Reproducibility}
Protocol, hypotheses, thresholds, the frozen sampler and the analysis plan were registered
before the first run, and every gate is logged with its measured values. The released
artefact holds the harness, all \Nruns\ run records, the \Nsurface-point surface, and the
fitting and bootstrap code. Training the grid took \Ngpuhours\ GPU-hours.

\section{The shape grid, as run}
\label{app:grid}
Every config in the active grid with its measured $T{=}16$ means (seeds
\Nseedlist, \Nitems\ items each from the zero-shot evaluation set eval\_zs). Iso-$N$ within a budget: the three budgets (A, B, C) sit
at 20, 50 and 125\,M non-embedding parameters, and the five shapes per budget span the
aspect-ratio range at fixed $N = 12\,\dc{d}\,\wc{w}^{2}$.

\begin{center}
\small
\begin{tabular}{llrrrrrrrr}
\toprule
config & budget & \wc{$w$} & \dc{$d$} & heads & $N$ (M) & val loss & WER (\%) & SIM-o & Degen (\%) \\
\midrule
A1 & A & 640 & 4 & 10 & 19.7 & 4.638 & 53.9 & 0.327 & 0.67 \\
A2 & A & 448 & 8 & 7 & 19.3 & 4.582 & 28.4 & 0.348 & 0.08 \\
A3 & A & 384 & 12 & 6 & 21.2 & 4.566 & 15.9 & 0.360 & 0.17 \\
A4 & A & 320 & 18 & 5 & 22.1 & 4.574 & 13.6 & 0.359 & 0.00 \\
A5 & A & 256 & 24 & 4 & 18.9 & 4.611 & 15.3 & 0.351 & 0.17 \\
B1 & B & 832 & 6 & 13 & 49.8 & 4.551 & 27.8 & 0.361 & 0.08 \\
B2 & B & 640 & 10 & 10 & 49.2 & 4.505 & 16.0 & 0.381 & 0.00 \\
B3 & B & 512 & 16 & 8 & 50.3 & 4.497 & 11.9 & 0.388 & 0.08 \\
B4 & B & 448 & 22 & 7 & 53.0 & 4.511 & 12.0 & 0.382 & 0.08 \\
B5 & B & 384 & 30 & 6 & 53.1 & 4.520 & 11.5 & 0.375 & 0.08 \\
C1 & C & 1152 & 8 & 18 & 127.4 & 4.504 & 17.1 & 0.373 & 0.17 \\
C2 & C & 960 & 12 & 15 & 132.7 & 4.504 & 18.9 & 0.373 & 0.17 \\
C3 & C & 768 & 18 & 12 & 127.4 & 4.488 & 14.3 & 0.392 & 0.00 \\
C4 & C & 640 & 26 & 10 & 127.8 & 4.489 & 13.4 & 0.390 & 0.08 \\
C5 & C & 512 & 36 & 8 & 113.2 & 4.487 & 9.9 & 0.388 & 0.00 \\
\bottomrule
\end{tabular}

\end{center}

\begin{table}[h]
  \centering\small
  \caption{\emph{The complete pre-registration ledger.} Every hypothesis declared before the
  first run (H-D: main grid, v1.0; H-E: extensions; H-S: speaker identity), with its verdict, including the four that failed and the two that split. We
  report the failures beside the successes because a ledger that shows only its winners
  cannot be used to judge the winners. \textsc{discordant} is a declared outcome class, used
  where the two pre-specified lenses (primary and secondary tests) disagree, and is not a synonym for refuted.}
  \footnotesize
  \begin{tabular}{lll}
    \toprule
    id & claim & verdict \\
    \midrule
    H-D1 & fixed-$T$ anisotropy $\Delta\rho>0$ & not supported \\
    H-D2 & $\Delta\tau>0$, metric-selective steps & supported$^{\S}$ \\
    H-D3 & depth--step exchange rate $\kappa$ & not supported \\
    H-D4 & small budgets predict the largest & not supported \\
    H-E1a & $T^{*}_{\mathrm{WER}}>16$ & supported \\
    H-E1b & extended $\tau$ inside the v1.0 CI & \textbf{refuted} \\
    H-E2 & iso-latency allocation is depth-first & \textbf{refuted} \\
    H-E3 & coarse-level NFE beats fine & supported \\
    H-E4a & scale persistence, four budgets & supported \\
    H-E4b & predicted vs observed within 15\,\% & \textbf{refuted} \\
    H-E5 & survives $3\times$ training compute & supported \\
    H-S1 & longer prompt context buys identity & discordant \\
    H-S2 & search buys identity & supported \\
    H-S3 & rate-matched length buys identity & discordant \\
    H-S4 & contrastive guidance buys identity & \textbf{refuted} \\
    \midrule
    \multicolumn{3}{l}{\footnotesize $^{\S}$supported as declared; see \S\ref{sec:scope} for
    the coordinate, range and fit conditions that bound it.} \\
    \bottomrule
  \end{tabular}
  \label{tab:prereg}
\end{table}

\begin{table}[h]
  \centering\small
  \caption{\emph{Inference-time interventions that do not buy identity.} Speaker-contrastive
  guidance trades intelligibility for similarity and degrades monotonically in $\gamma$;
  per-item rate matching gives a small mean gain that is not reliable item-by-item.}
  \begin{tabular}{llccc}
    \toprule
    intervention & setting & $\Delta$ SIM-o & 95\% CI & $\Delta$ WER (pts) \\
    \midrule
    contrastive guidance & $\gamma = 0.5$ & \NgamLosim & \NgamLoci & \NgamLower \\
                         & $\gamma = 1.0$ & \NgamMidsim & \NgamMidci & \NgamMidwer \\
                         & $\gamma = 2.0$ & \NgamHisim & \NgamHici & \NgamHiwer \\
    \midrule
    rate matching & per-item duration & \NrateGainSim & \NrateCI & \NrateWER \\
    \multicolumn{5}{l}{\footnotesize \quad \NrateRuns/15 runs positive, but a per-item sign
    test gives $p = \NrateP$} \\
    \bottomrule
  \end{tabular}
  \label{tab:negative}
\end{table}
\begin{table}[h]
  \centering\small
  \caption{\emph{Longer prompts do not buy identity, and destroy intelligibility.} Prompt
  length swept at inference on 15 runs of the \Nbudgetc\ budget, 397 items, every arm scored against
  the same fixed 3\,s reference. Training used 3\,s prompts only, so the 6 and 9\,s arms
  test extrapolation. There the output stays fluent and speaker-consistent, but it decouples
  from the conditioning text.}
  \begin{tabular}{lcc}
    \toprule
    prompt & SIM-o & WER \\
    \midrule
    1.5\,s & \NctxAsim & \NctxAwer \\
    3\,s \emph{(training length)} & \textbf{\NctxBsim} & \textbf{\NctxBwer} \\
    6\,s & \NctxCsim & \NctxCwer \\
    9\,s & \NctxDsim & \NctxDwer \\
    \bottomrule
  \end{tabular}
  \label{tab:context}
\end{table}

\begin{table}[h]
  \centering\small
  \caption{\emph{Where the refinement passes are spent matters as much as how many.} Three
  frozen per-level schedules (steps per codebook level, coarsest first) at matched total NFE${}=32$ on the \Nbudgetc\ budget, 15 runs,
  400 items. Spending every pass on the coarsest codebook level rather than the finest cuts
  WER by \Nfivepointtwo$\times$, and costs identity and naturalness against a uniform split.}
  \begin{tabular}{lcccc}
    \toprule
    schedule & WER & SIM-o & UTMOS & degenerate \\
    \midrule
    coarse \texttt{[25,1,\dots,1]} & \NallocCwer & \NallocCsim & \NallocCut & \NallocCdeg\% \\
    uniform \texttt{[4,\dots,4]}   & \NallocUwer & \NallocUsim & \NallocUut & \NallocUdeg\% \\
    fine \texttt{[1,\dots,1,25]}   & \NallocFwer & \NallocFsim & \NallocFut & \NallocFdeg\% \\
    \bottomrule
  \end{tabular}
  \label{tab:alloc}
\end{table}

\begin{table}[h]
  \centering
  \caption{\emph{Fitted exponents} on the \Nsurface-point surface ($M_{\mathrm{sep}}$ for
  $\alpha,\beta,\tau$; $M_{\mathrm{sub}}$ for $\kappa$), with 95\% run-level bootstrap
  intervals on the step exponents. The primary result is $\Delta\tau = \Ndtau$, CI \Ndtauci,
  which excludes zero. $\kappa$ sits on its bound in both rows, and the substitution form is
  worse than the separable one by $\Delta\mathrm{AICc}_{\mathrm{sub-sep}} \gg +4$, so the
  depth--step exchange rate has no finite estimate. The negative
  $\Delta\mathrm{AICc}_{\mathrm{full}-N}$ ($M_{\mathrm{full}}$, the separable form without its $T$ term, fitted at $T{=}16$, against a single power law in $N$) says that shape matters at fixed $N$.}
  \small
  \begin{tabular}{lcccccc}
    \toprule
    Metric & $\alpha$ (\wc{width}) & $\beta$ (\dc{depth}) & $\tau$ (\tc{steps}) & $\kappa$ &
    $\Delta$AICc$_{\mathrm{sub-sep}}$ & $\Delta$AICc$_{\mathrm{full}-N}$\\
    \midrule
    WER   & \NalphaBWER & \NbetaBWER & \textbf{\NtauWER}\ \Ntauwerci & \Nkappawer$^{\dagger}$ & $+\NaiccSSWER$ & $\NaiccFNWER$\\
    SIM-o & \NalphaBSIM & \NbetaBSIM & \textbf{\NtauSIM}\ \Ntausimci & \Nkappasim$^{\dagger}$ & $+\NaiccSSSIM$ & $\NaiccFNSIM$\\
    \midrule
    \multicolumn{7}{l}{\footnotesize $^{\dagger}$at the fitted bound, i.e.\ not identified; $\Delta\tau=\mathbf{\Ndtau}$, CI \Ndtauci.}\\
    \bottomrule
  \end{tabular}
  \label{tab:exponents}
\end{table}

\begin{table}[h]
  \centering\small
  \caption{\emph{$\Delta\tau$ survives variation in scale, training compute and the
  measurement stack.} Each row refits the surface with one factor changed: a fourth parameter
  budget, $3\times$ training compute, a different ASR, and an unbounded reparameterisation of
  the fitted amplitudes. Every interval excludes zero.}
  \begin{tabular}{llcc}
    \toprule
    variation & what changes & $\Delta\tau$ & 95\% CI \\
    \midrule
    none (\Nsurface\ points, \Nruns\ runs) & n/a & \Ndtau & \Ndtauci \\
    fourth budget, \NmaxNfour\,M & 10 further runs & \Ndtaufour & \Ndtaufourci \\
    $3\times$ training compute & 3 configs at 90k steps & \Ndtaunineok & \Ndtaunineokci \\
    ASR swapped & Whisper-medium.en & \Ndtauasr & \NdtauAsrCI \\
    unbounded amplitudes & $\log A, \log B, \log C$ free & \Ndtaulogamp & \NdtauLogCI \\
    \bottomrule
  \end{tabular}
  \label{tab:robust}
\end{table}

\begin{figure}[h]
  \centering
  \includegraphics[width=.62\linewidth]{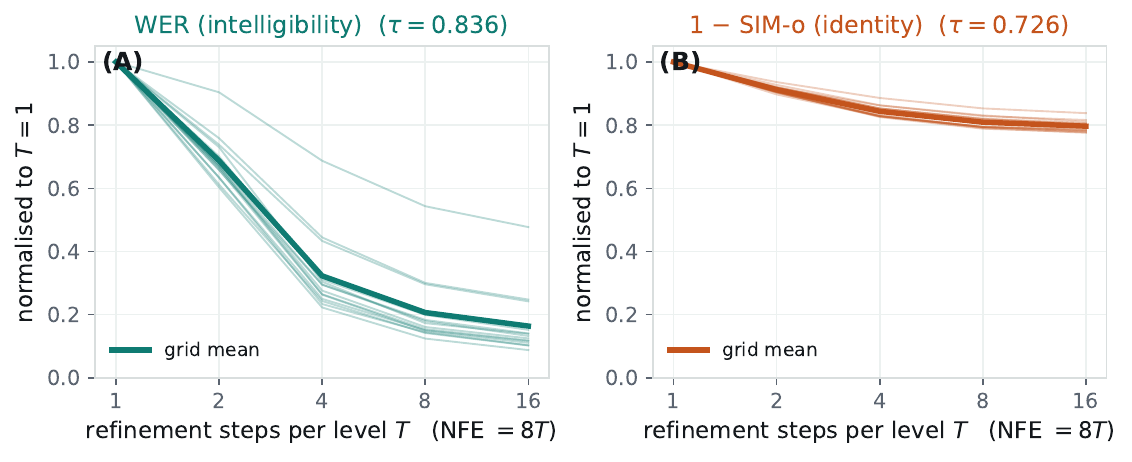}
  \caption{\emph{Test-time scaling is metric-selective.} Error versus
  \tc{refinement steps} for all \Nconfigs\ configs, each normalised to its own
  $\tc{T}{=}1$ value; thick lines are grid means. \textbf{(A)}~WER error falls \NabsWERgain\ in
  absolute terms over $\tc{T}{=}1\to16$ and is still falling at $\tc{T}{=}16$.
  \textbf{(B)}~SIM-o error falls \NabsSIMgain\ over the same range, \NmagAsym\ less.
  Same models, same frozen sampler, same items; only
  \tc{$T$} differs.}
  \label{fig:main}
\end{figure}

\FloatBarrier
\section{Additional figures}
\label{app:figs}


\begin{figure}[h]
  \centering
  \includegraphics[width=.86\linewidth]{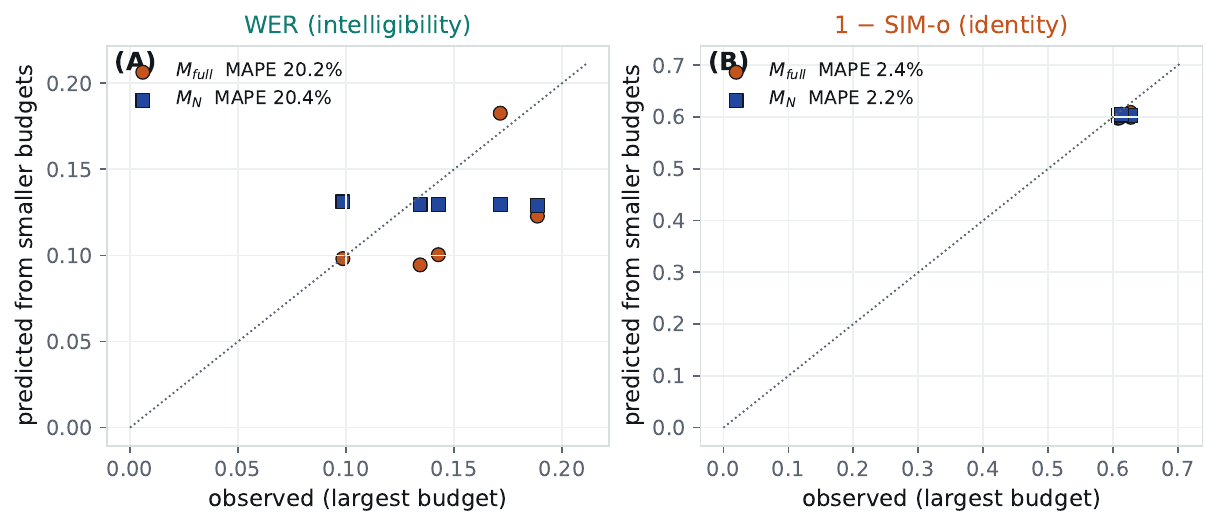}
  \caption{\emph{H-D4 extrapolation.} The two smaller budgets are fitted and the largest
  budget predicted, under $M_{\mathrm{full}}$ and the $N$-only model. SIM-o extrapolates
  (mean absolute percentage error, MAPE, \NmapeSIM\,\% vs \NmapeNSIM\,\%); WER does not (\NmapeWER\,\% vs \NmapeNWER\,\%).}
  \label{fig:extrap}
\end{figure}

\FloatBarrier
\section{Refinement-budget allocation across codebook levels}
\label{app:alloc}
Three frozen per-level step schedules at matched total NFE${}=32$ on the \Nbudgetc\ budget
(15 runs, 400 items). Paired item-level bootstrap, \Nbootreps\ replicates, pairing on the
eval item so item difficulty cancels. Coarse spends all passes on codebook level 0 (the coarsest), fine on
level 7 (the finest), and uniform splits evenly.

\begin{center}\small
\begin{tabular}{lcccc}
\toprule
schedule & WER & SIM-o & UTMOS & degenerate \\
\midrule
coarse \texttt{[25,1,...,1]}  & \NnfeCoarseWER  & \NnfeCoarseSIM  & \NnfeCoarseUT  & \NnfeCoarseDegen\% \\
uniform \texttt{[4,...,4]}    & \NnfeUniformWER & \NnfeUniformSIM & \NnfeUniformUT & \NnfeUniformDegen\% \\
fine \texttt{[1,...,1,25]}    & \NnfeFineWER    & \NnfeFineSIM    & \NnfeFineUT    & \NnfeFineDegen\% \\
\bottomrule
\end{tabular}
\end{center}

\noindent Allocation is metric-selective in the same way the step \emph{count} is: coarse
wins WER but loses identity and naturalness against uniform, both with intervals excluding
zero. It is a trade, not a free win.

\paragraph{A significance-based saturation contrast.} Defining saturation as the smallest
$T$ whose paired-bootstrap interval for $\mathrm{err}(T)-\mathrm{err}(64)$ covers zero
gives $T^{*}_{\mathrm{WER}}=32$ against $T^{*}_{\mathrm{SIM}}=16$ on the extended grid.
We report this only here and do not build on it: roughly half the contrast is a precision
artefact, since identity's detection floor is ${\approx}1.8\times$ looser relative to its own
dynamic range than intelligibility's. The raw-scale $T^{*}$ (95\% of the
$T{=}16$ \emph{value}) is not affine-invariant and must never be compared across metrics:
the same 5\% band is 0.67\% of WER's range but 11.96\% of SIM-o's. On the invariant
fraction-of-total-gain definition, $T^{*}=16$ for both metrics.

\section{Fit diagnostics and exploratory analyses}
\label{app:diag}
\textbf{Parameters at bounds.} The width amplitude $A$ saturates the pre-registered
\emph{upper} bound $A \le 10$ in every fit (Part A, the $T{=}16$ width--depth fit: 10.000 for both metrics; Part B, the full surface fit: 9.999
for WER, 10.000 for SIM-o), so only the product $A\,w^{-\alpha}$ is identified and
$\alpha$ is not; the bound, not the data, is binding, which is a limitation of the
pre-registered parameterisation worth fixing in any follow-up. $\kappa$ reaches its upper
bound in every one of the \Nbootreps\ bootstrap replicates of $M_{\mathrm{sub}}$. The
irreducible term $E$ reaches 0 for WER under $M_{\mathrm{sep}}$. The depth amplitude $B$ is
interior throughout, so $\beta$ is identified.

\textbf{Exploratory: the depth exponent separates the capabilities.}
$\beta_{\mathrm{WER}} = \NbetaBWER$ against $\beta_{\mathrm{SIM}} = \NbetaBSIM$, a factor
of 3.7: the asymmetry H-D1 sought, in the parameter the data pin down. Part-B $\rho$
values point the pre-registered way ($\Delta\rho = +0.890$) but inherit the $A$-cap
problem; the pre-registered test is Part A, so this is exploratory only.

\textbf{Weighting.} Points are weighted $1/\mathrm{SE}^2$ with SE taken over seeds, floored
at a quarter of the pooled seed SD so that a chance agreement between two seeds cannot
dominate the fit; the number of floored points is recorded in the released
\texttt{fits.json}.

\textbf{Exploratory (not part of any declared claim).} An unweighted refit of both parts is
released alongside the weighted fit, as is a fit that additionally includes the finished
runs of a stretch third seed. Both reproduce the sign and significance of $\Delta\tau$.

\section*{Use of Generative AI Disclosure}
Generative AI was used as a coding assistant for the experimental harness and for language
refinement of author-written text. The research design, the pre-registered hypotheses and
thresholds, the analysis plan, and all scientific claims are the authors' own; the authors
verified the code and the numbers and take full responsibility for this paper.

\end{document}